\documentclass{article}

\usepackage{amsmath}
\usepackage{amssymb}
\usepackage{amsthm}
\usepackage{placeins}
\usepackage{wrapfig}
\usepackage{subfigure}
\usepackage{nicefrac}

\usepackage{hyperref}

\newcommand{\E}[2]{\operatorname{\mathbb{E}}_{#1}\left[#2\right]}

\newcommand{\density}{p}

\newcommand{\entropy}{\mathcal{H}}
\newcommand{\ent}{\mathcal{H}}

\newcommand{\mutualinfo}{\mathcal{I}}

\newcommand{\voidarg}{{\,\cdot\,}}

\newcommand{\sspace}{\mathcal{S}}
\newcommand{\aspace}{\mathcal{A}}
\newcommand{\gspace}{\mathcal{G}}

\newcommand{\state}{\mathbf{s}}

\newcommand{\st}{{\state_t}}

\newcommand{\stp}{{\state_{t+1}}}

\newcommand{\pdyn}{\density}

\newcommand{\action}{\mathbf{a}}
\newcommand{\goal}{\mathbf{g}}

\newcommand{\at}{{\action_t}}
\newcommand{\gt}{{\goal_t}}

\newcommand{\reward}{r}

\newcommand{\extreward}{f}

\newcommand{\fmin}{f_\mathrm{min}}
\newcommand{\fmax}{f_\mathrm{max}}

\newcommand{\V}{V}

\newcommand{\Q}{Q}

\newcommand{\policy}{\pi}

\newcommand{\params}{\theta}

\newcommand{\pparams}{{\phi}}   
\newcommand{\vparams}{{\psi}}   

\newcommand{\apolicy}{{\policy_\action}}
\newcommand{\gpolicy}{{\policy_\goal}}
\newcommand{\agpolicy}{{\policy_{\action\goal}}}

\newcommand{\reals}{\mathbb{R}}

\newcommand{\discount}{\gamma}

\newcommand{\aref}[1]{\hyperref[#1]{Appendix~\ref*{#1}}}

\usepackage{microtype}
\usepackage{graphicx}
\usepackage{subfigure}
\usepackage{booktabs} %
\usepackage{mathtools,amssymb,lipsum, nccmath}

\usepackage{tikz} 
\usetikzlibrary{shapes,arrows} 

\usepackage[accepted]{icml2018}

\icmltitlerunning{Hierarchical Soft Actor-Critic}

\begin{document}

\twocolumn[
\icmltitle{{\LARGE Hierarchical Soft Actor-Critic:} \\ Adversarial Exploration via Mutual Information Optimization }

\icmlsetsymbol{equal}{*}

\begin{icmlauthorlist}
\icmlauthor{Ari Azarafrooz}{Cylance}
\icmlauthor{John Brock}{Cylance}
\end{icmlauthorlist}

\icmlaffiliation{Cylance}{Department of Research and Intelligence, Blackberry Cylance, Irvine, USA}
\icmlcorrespondingauthor{Ari Azarafrooz}{mazarafrooz@cylance.com}

\icmlkeywords{reinforcement learning, control as inference}

\vskip 0.3in
]

\printAffiliationsAndNotice{}  %

\begin{abstract}
We describe a novel extension of soft actor-critics for hierarchical Deep Q-Networks (HDQN) architectures using mutual information metric. The proposed extension provides a suitable framework for encouraging explorations in such hierarchical networks. A natural utilization of this framework is an adversarial setting, where meta-controller and controller play minimax over the mutual information objective but cooperate on maximizing expected rewards. 

\end{abstract}

\section{Introduction}
Learning in environments with \emph{sparse} reward feedback has been one of the major challenges for reinforcement learning frameworks. Exploration is one of the key components in the design of data-efficient RL for such complex environments. 

One recent proposed approach for inducing explorations is via hierarchical value functions (HDQN) [1].  Such hierarchical organization provides efficient explorations by breaking the problem into various levels of spatiotemporal abstractions. 
More specifically,  HDQN represents the problem of expected reward maximization via two Q-Networks stacked in a hierarchical way. A high-level meta-controller maximizes the external rewards by proposing \emph{explicit} sub-goals for a low-level controller to achieve. 
A further advantage of such method is that the explicit sub-goals can be defined to constrain the exploration space for complex environments even further, for example, by defining goals in the space of the entities and relations, via domain-specific knowledge of the environment or learning them beforehand.

We show that one can design even more data-efficient hierarchical RL algorithms by reframing the objective of HDQN at each level of abstractions, as a maximum entropy reinforcement learning  (ME-RL) and utilizing soft-actor critic (SAC) method of [2]. The maximum entropy term in ME-RL formulation is to incentivize the controller to explore more. However, one important consideration is that for a meaningful exploration for such hierarchical architectures one should account for the \emph{interplay} of explorations on various levels of abstractions. How should we make sure that the explorations of a higher level controller don't interfere with the meaningful explorations of lower level one and vise-vera? As the main contribution of this paper,  we will extend the ME-RL framework to that of the mutual information RL (MI-RL) framework. This formulation not only helps to decouple the explorations of agents at various level of hierarchy but also can be used as a proper setup for adversarial explorations where meta-controller and controller play minimax over the mutual information objective but cooperate on maximizing expected rewards. 

\vspace{-1mm}
\section{Preliminaries}
\label{sec:preliminaries}

\vspace{-1mm}
\subsection{Notation}
 Consider a Markov decision process (MDP), represented by the tuple $(\sspace, \aspace, T, \extreward)$, where $\sspace$ is the state space, $\aspace$ is the action space and $T:\sspace \times \aspace \rightarrow  \sspace$ is assumed to be an unknown transition function. The environment emits a bounded reward $\extreward: \sspace \times \aspace \rightarrow  [\fmin,\fmax]$ on each transition.  We will use $\rho_\policy(\st)$ and $\rho_\policy(\st,\at)$ to denote the state and state-action marginals of the trajectory distribution induced by a policy $\policy(\at|\st)$. 

\subsection{Hierarchical Reinforcement Learning}
 The set of actions in HDQN are $\{\aspace,\gspace\}$, with $\gspace$ being \emph{temporally} more \emph{abstract} than $\aspace$. The expected sum of environment rewards is maximized by the cooperation of a controller and meta-controller. Meta-controller takes high-level actions $\goal$ and delegates it to the controller. Since $\goal$ serves as the subgoal for the controller to achieve, it is referred to as subgoal.  The controller gets rewarded $\reward$ for achieving subgoals. For example, assuming a cooperative behavior it receives a reward of 1 when $goal$ is achieved and 0 otherwise.

In standard HDQN, a meta-controller maximizes $\sum_t \E{(\st,\gt)\sim\rho_\gpolicy}{\extreward(\st,\gt)}$. 
Controller maximizes the expected sum of internal rewards conditioned on the provided subgoals: $\sum_t \E{(\st,\at)\sim\rho_\agpolicy}{\reward(\st,\at|\gt)}$. 

\section{Maximum Entropy Hierarchical RL}
The ME-RL can be easily adopted for both meta-controller and controller. ME-RL favors stochastic policies by augmenting the objective with the expected entropy of the policy over  $\rho_\apolicy(\st)$ for the meta-controller
and  $\rho_\agpolicy(\st)$ for the controller as is shown in the following formulations: 

\begin{align}
\label{eq:meta_controller_maxent_objective}
\medmath{J(\gpolicy) =\sum_{t=0}^{T} \E{(\st, \gt) \sim \rho_\gpolicy}{\reward(\st,\gt) + \alpha\ent(\gpolicy(\voidarg|\st))}}
\end{align}

\begin{align}
\label{eq:controller_maxent_objective}
\medmath{J(\agpolicy) =\sum_{t=0}^{T} \E{(\st, \at) \sim \rho_\agpolicy}{\reward(\st,\at|\gt) + \alpha\ent(\agpolicy(\voidarg|\st,\gt))}}
\end{align}

The temperature parameter $\alpha$ determines the relative importance of the entropy term against the reward. Form ME-entropy framework it is not clear how the introduced stochasticity for encouraging exploration of abstract actions $\goal$ interact with stochasticity required for the explorations of atomic actions $\action$.  In the next section, we show how the controller objective Eq. \ref{eq:controller_maxent_objective} can be modified for introducing more meaningful stochasticity.

\section{Mutual Information Hierarchical RL}

In the Mutual Information Hierarchical RL setup we set the meta-controller objective to simply follow Eq. \ref{eq:meta_controller_maxent_objective}. However, we note that controller requires to encourage explorations with respect to all available action spaces (both atomic and abstract) $\{\aspace,\gspace\}$. A natural objective function is \emph{mutual information} $\mutualinfo(\action;\goal|\state)=\entropy(\apolicy(\voidarg|\state))-\entropy(\agpolicy(\voidarg|\state,\goal))$. $\mutualinfo(\action;\goal | \state)=0$ if and only if $\action$ and $\goal$ are independent random variables given state $\state$. At the other extreme, if $\action$ is a deterministic function of $\goal$, it reduces to $\entropy(\apolicy(\voidarg|\state))$. We suggest to replace Eq. \ref{eq:controller_maxent_objective} with Eq. \ref{eq:maxinf_objective}.

\begin{align}
\label{eq:maxinf_objective}
\medmath{J(\agpolicy) = \sum_{t=0}^{T} \E{(\st, \at) \sim \rho_\agpolicy}{\reward(\gt, \st, \at) - \alpha\mutualinfo(\at;\gt|\st)}}
\end{align}

\emph{This formulation is set up to encourage the controller policy $\agpolicy$ toward actions that are independent of goals}. 

\subsection{Adversarial MI- Hierarchical RL}
The mutual information framework for encouraging explorations allows for the meta-controller to participate in the optimization of $\mutualinfo(\action;\goal | \state)$ as well. For example, it yields to a \emph{minimax one}, when meta-controller's objective is to maximize $\mutualinfo(\action;\goal | \state)$. 

If we assume an information channel between $\goal$ and $\action$, the controller's objective is to minimize this information rate while meta-controller's is to increase this rate according to to the minimax optimization:

\begin{align}
\label{eq:adv-channel}
\medmath{\min_{{\agpolicy}} \max_{\gpolicy} ~ \entropy(\apolicy(\voidarg|\state))-\entropy(\agpolicy(\voidarg|\state,\goal))} 
\end{align}

 This is consistent with the game-theoretic incentive of its rewards, \emph{as it encourages the controller to follow its delegated subgoals \emph{deterministically}}. In other words, a consistent game theoretic set up is when meta-controller and controller cooperate on maximizing rewards but play minimax over the mutual information objective. Cooptation over rewards is consistent with adversarial setup over exploration.

\section{Hierarchical Soft Actor-Critic} 
Our proposed RL algorithm is based on the previous formulation of off-policy soft-actor critic of [2]. For the sake of brevity, we describe only the main part of the departures from soft actor-critic algorithm of [2]. We do however leave the theoretical proof of convergence of the algorithm to optimal policies for future works.  We only focus on the description of the mutual information variant of a policy iteration method for the controller as the derivation of the ME-RL version using [2] is straightforward. 

We first outline the update mechanisms for the controller.  Depending on the update mechanisms for the meta-controller, we can have two types of soft-actor critics. Therefore, we describe the meta-controller update mechanisms in a separate section.

\subsection{Controller}

In the policy evaluation step, we wish to compute the value of a policy $\agpolicy$ according to the mutual information objective in~\autoref{eq:maxinf_objective}. For fixed policies ($\gpolicy,\agpolicy)$, the soft $Q$-value can be computed iteratively, starting from any function $Q: \sspace\times \aspace \rightarrow \reals$ and repeatedly applying a modified Bellman backup operator $\mathcal{T}^\policy$ given by
\begin{align}
\label{eq:soft_bellman_backup_op}
\medmath{\mathcal{T}^\policy Q(\gt, \st, \at) \triangleq  \reward(\gt, \st, \at) + \discount \E{\stp \sim \pdyn, g_{t+1} \sim \gpolicy}{V_1(\stp)}}
\end{align}
where
\begin{align}
\medmath{V_1(\st) = \E{\at\sim\agpolicy}{\Q(\gt, \st, \at)} - \mutualinfo(\at;\gt|\st)}
\label{eq:soft_value_function}
\end{align}
is the MI-soft state value function for the controller

We use function approximators for both the Q-functions and the policies.
We will consider a parameterized soft Q-function $\Q^\params_1(\gt, \st, \at)$ for the controller with tractable policy $\agpolicy^\pparams(\at|\gt, \st)$.The parameters of these networks are $\params_1$, and $\pparams$.
For the meta-controller, will consider a parameterized soft Q-function $\Q^\params_2(\gt, \st)$ with tractable policy $\gpolicy^\vparams(\gt|\st)$.The parameters of these networks are $\params_2$, and $\vparams$.

 $\Q^\params_1$ can be trained to minimize the following soft Bellman residual:
\begin{align}
\medmath{J_\Q(\params_1) = \E{(\gt, \st, \at)\sim\mathcal{D}}{\frac{1}{2}\left(\Q^\params_1(\gt, \st, \at) - \hat \Q^\params_1(\gt, \st, \at)\right)^2}}
\label{eq:q_cost}
\end{align}
with 
\begin{align}
\medmath{\hat \Q^\params_1(\gt, \st, \at)  = \reward(\gt, \st, \at) + \discount \E{\stp\sim\pdyn, g_{t+1} \sim \gpolicy}{\V_1(\stp)}}
\end{align}, where $\reward$ is the internal reward. It is 1 when goal is achieved and 0 otherwise. $\mathcal{D}$ is the distribution of previously sampled states and actions, or a replay buffer. 

We don't use any other separate function approximator (such as a second Q-Network to address maximization bias)

Considering the explicit and discrete nature of the subgoals $\goal$, in order to maintain differentiability, we will restrict the policies to a special parameterized family of distributions known as GUMBEL-SOFTMAX [3].
Since the target density (Q-function) is represented by a neural network and hence is differentiable, we use reparameterization trick using a neural network that returns the parameters of GUMBEL-SOFTMAX.

Controller policy gets updated according to: 
\begin{align}
J_\agpolicy(\pparams) = \arg\underset{\agpolicy^\pparams \in \Pi}{\min} \entropy(\E{\gpolicy^\vparams}{\agpolicy^\pparams( \voidarg|\st,\gt)})\notag\\
-\entropy(\agpolicy^\pparams(\voidarg|\st,\gt)) -\E{\agpolicy^\pparams}{Q^\params_1(\st, \voidarg)}
\label{eq:policy_objective}
\end{align}
Note that we rewrote the mutual information in terms of only ($\gpolicy,\agpolicy)$.

\subsection{Meta-controller}

\subsubsection{Hierarchical Mutual Information-soft actor-critic}

The meta-controller policy gets updated according to the ME soft-actor critic of [2] with the following MI-soft state value function  
\begin{align}
\medmath{V_2(\st) = \E{\gt\sim\gpolicy}{\Q^\params_2(\gt, \st)} + \entropy(\gt|\st)}
\label{eq:soft_value_function_meta}
\end{align}
is the MI-soft state value function for the controller

 $\Q^\params_2$ can be trained to minimize the following soft Bellman residual:
\begin{align}
J_\Q(\params_2) = \E{(\st, \at)\sim\mathcal{D}}{\frac{1}{2}\left(\Q^\params_2(\st, \at) - \hat \Q^\params_2(\st, \at)\right)^2},
\label{eq:q_cost}
\end{align}
with 
\begin{align}
\hat \Q^\params_2(\st, \at) = \extreward(\st, \at) + \discount \E{\stp\sim\pdyn}{\V_2(\stp)},
\end{align}

where $\extreward$ is the external reward.

\subsubsection{Adversarial Hierarchical Mutual Information-soft actor-critic}

Using Eq. \ref{eq:adv-channel} we update the meta-controller policy according to the following formulation:

\begin{align}
J_\gpolicy(\vparams) = \arg\underset{\gpolicy^\vparams \in \Pi}{\min} -\entropy(\E{\gpolicy^\vparams}{\agpolicy^\pparams( \voidarg|\st,\gt)})\notag\\
+\entropy(\agpolicy^\pparams(\voidarg|\st,\gt)) -\E{\gpolicy^\vparams}{Q^\params_2(\st, \voidarg)}
\label{eq:policy_objective}
\end{align}

 $\Q^\params_2$ can be trained to minimize the following soft Bellman residual:
\begin{align}
\medmath{J_\Q(\params_2) = \E{(\gt, \st)\sim\mathcal{D}}{\frac{1}{2}\left(\Q^\params_2(\gt, \st, \at) - \hat \Q^\params_2(\gt, \st, \at)\right)^2}}
\label{eq:q_cost}
\end{align}
with 
\begin{align}
\medmath{\hat \Q^\params_2(\gt, \st)  = \extreward(\gt, \at) + \discount \E{\stp\sim\pdyn}{\V_2(\stp)}}
\end{align}

where $\V_2$ is the MI-soft state value function of the following:
\begin{align}
\medmath{V_2(\st) = \E{\gt\sim\gpolicy}{\Q^\params_2(\gt, \st)} + \mutualinfo(\at;\gt|\st)}
\label{eq:soft_value_function}
\end{align}
is the MI-soft state value function for the controller

\begin{figure*}[h]
    \centering
    \subfigure[$n_g=6$]{
        \includegraphics[width=0.45\textwidth, trim={0 0 5mm 7.5mm}, clip]{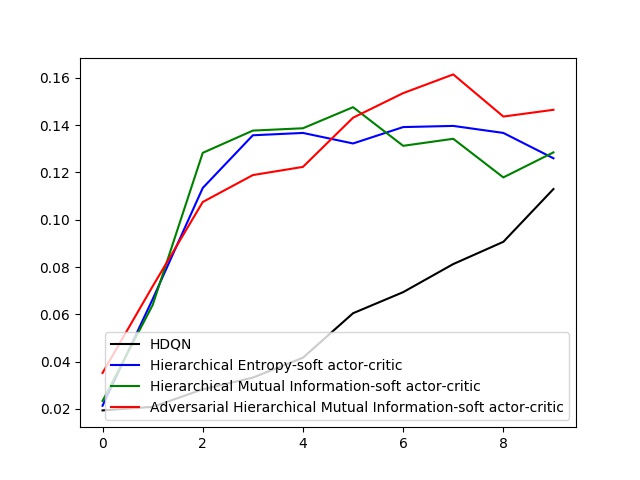}
    }
      \subfigure[$n_g=8$]{
        \includegraphics[width=0.45\textwidth, trim={0 0 5mm 7.5mm}, clip]{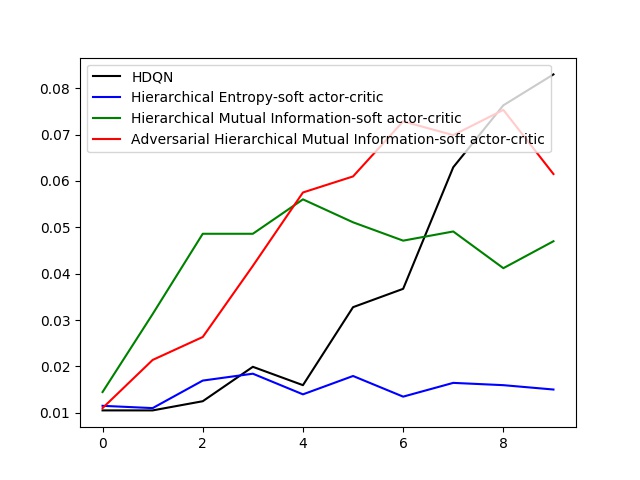}
    }

      \subfigure[$n_g=12$]{
        \includegraphics[width=0.45\textwidth, trim={0 0 5mm 7.5mm}, clip]{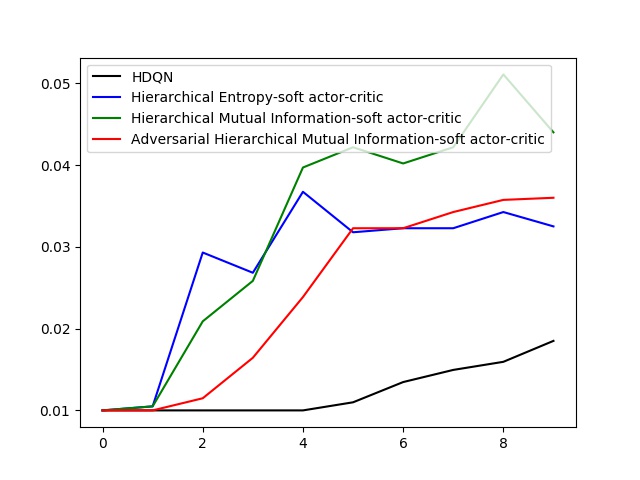}
    }
    \subfigure[$n_g=18$]{
        \includegraphics[width=0.45\textwidth, trim={0 0 5mm 7.5mm}, clip]{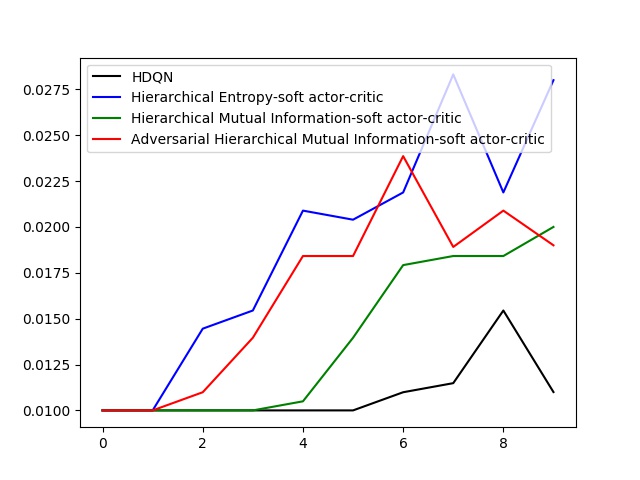}
        
    }
    \caption{\small  The effect of exploration strategy in hierarchical RL with explicit subgoals for various levels of feedback sparsity $n_g$. We consider each state as a possible goal for exploration.}
    \label{fig:training_curves_all}
 \end{figure*}

    

    

       
         
    

\section{Experiments}
We performed experiments on a discrete-state MDP with stochastic transitions. It is a stochastic decision process where the extrinsic reward depends on the history of visited states in addition to the current state. [1] used the same experiment to showcase the importance of exploration in such environments. In that, there are $n$ possible states and the agent always starts at $s_2$. The agent moves left deterministically when it chooses \emph{left} action; but the action \emph{right} only succeeds 50\% of the time, resulting in a left move otherwise. The terminal state is $s_1$ and the agent receives the reward of 1 when it first visits $s_{n_g}$ and then $s_1$. The reward for going to $s_1$ without visiting $s_{n_g}$ is 0.01. We can consider each state as a possible goal for exploration and use $n_g=|\gspace|$ as a control parameter to adjust the sparsity of the environment feedbacks.  We showed the performance of 'HDQN', 'Hierarchical Entropy-soft actor-critic', 'Hierarchical Mutual Information-soft actor-critic' and 'Adversarial Hierarchical Mutual Information-soft actor-critic' in such environment for various degree of feedback sparsity $n_g$. The results are for the average external reward of window size of 100, averaged over 20 random runs of the experiments.  We found out the most sensitive parameter to be the temperature of GUMBEL-SOFTMAX. It is known that there is a tradeoff between
small temperatures, where samples are close to one-hot but the variance of the gradients is large,
and large temperatures, where samples are smooth but the variance of the gradients is small. We set the temperature of 0.3 in our experiments. All the neural networks have 2 hidden layers of size 256. For all Q-Networks except the HDQN, we add the dropout 0.2 to every layer. Activation functions are ReLU except for the last layer of the Policy nets which are Softmax.

The results of the experiments are shown in Fig. \ref{fig:training_curves_all}.  Higher $n_g$ implies sparser feedbacks. The results confirm that a mutual-information framework is suitable for encouraging explorations in Hierarchical RL. 
Also, it can be seen that the Adversarial Hierarchical Mutual Information-soft actor-critic consistently performs well across various levels of feedback sparsity $n_g$. 

\section{Conclusion}

A novel Mutual Information RL framework for encouraging further explorations in hierarchical Q-Network with explicit subgoals is introduced. The proposed framework is suitable for introducing adversarial exploration in such architectures. Simulation over various levels of feedback sparsity using discrete-state MDP generated data, shows the practicality of such frameworks using Hierarchical soft-actor critics. While these algorithms should still be tested against real-world data, the proposed framework provides a novel direction for encouraging explorations in hierarchical RL by minimax formulation of mutual information between stochasticity of abstract and atomic actions. 

\section{References}
[1] T. D. Kulkarni, K. R. Narasimhan, J. B. Tenenbaum and A. Saeedi. Hierarchical Deep Reinforcement Learning: Integrating Temporal Abstraction and Intrinsic Motivation. \emph{NeuroIPS2017}

[2] T. Haarnoja,  A. Zhou, P. Abbeel and S. Levine. Soft Actor-Critic: Off-Policy Maximum Entropy Deep Reinforcement Learning with a Stochastic Actor, \emph{ICML 2017}.

[3] E. Jang, S. Gu and B. Poole. Categorical Reparameterization With Gumble-Softmax,  \emph{ICLR 2017}.

\end{document}